\documentclass{article}
\usepackage[utf8]{inputenc}
\usepackage[letterpaper,margin=1.2in]{geometry}
\usepackage[parfill]{parskip}
\usepackage{xr-hyper,refcount}
\usepackage{graphicx}
\usepackage[utf8]{inputenc} 
\usepackage[T1]{fontenc}    

\usepackage[hyphens, spaces, obeyspaces]{url}     
\usepackage{hyperref}

\hypersetup{
    colorlinks=true,
    linkcolor=blue,
    filecolor=magenta,      
    urlcolor=blue,
}

\usepackage{booktabs}       
\usepackage{amsfonts}       
\usepackage{nicefrac}       
\usepackage{microtype}      
\usepackage[parfill]{parskip}
\usepackage{algorithm,algorithmic}
\usepackage{amsmath,amsthm,amssymb}
\usepackage{dsfont}
\usepackage{mathtools}
\usepackage{cases}
\usepackage{comment}
\usepackage{algorithm,algorithmic}
\usepackage{color}
\usepackage{appendix}
\usepackage{xspace}

\usepackage[english]{babel}
\usepackage{authblk}

\usepackage{libertine}
\usepackage[T1]{fontenc}

\usepackage[round]{natbib}

\usepackage{hyperref}       %
\usepackage{url}            %
\usepackage{booktabs}       %
\usepackage{amsfonts}       %
\usepackage{nicefrac}       %
\usepackage{microtype}      %
\usepackage{xcolor}         %

\usepackage{amsmath,amssymb}
\usepackage{booktabs}
\usepackage{url}
\usepackage{caption}
\usepackage{subcaption}

\title{Business Metric-Aware Forecasting for Inventory Management}
\date{}

\author{Helen Zhou$^1$, Sercan \"{O}. Ar{\i}k$^2$, Jingtao Wang$^2$\\
$^1$Carnegie Mellon University\\
$^2$Google\\
\text{
\href{mailto:hlzhou@andrew.cmu.edu}{hlzhou}@cmu.edu,
\{\href{mailto:soarik@google.com}{soarik},
\href{mailto:jingtaow@google.com}{jingtaow}\}@google.com}}

\usepackage{bibentry}

\begin{document}

\maketitle

\begin{abstract}
Time-series forecasts play a critical role in business planning.
However, 
forecasters
typically 
optimize objectives that are agnostic to downstream business goals and thus can produce forecasts misaligned with business preferences. 
In this work,
we demonstrate that optimization of conventional forecasting metrics can often lead to sub-optimal downstream business performance. 
Focusing on
the inventory management setting,
we derive an efficient procedure for computing and optimizing proxies of common downstream business metrics in an end-to-end differentiable manner. 
We explore a wide range of plausible cost trade-off scenarios, and empirically demonstrate that
end-to-end optimization often outperforms optimization of standard business-agnostic forecasting metrics (by up to 45.7\% for a simple scaling model, and up to 54.0\% for an LSTM encoder-decoder model).
Finally, we discuss how
our findings 
could benefit other
business contexts.
\end{abstract}

\section{Introduction}
Time-series forecasting is an essential component of decision-making 
and planning. 
In industries ranging from 
healthcare~\citep{jones2009multivariate,influenza_forecasting,cheng2021unpacking}, 
to finance \citep{thomas2000survey,elliott2016forecasting}, 
to energy \citep{AHMED2020109792,donti2021machine},
businesses
leverage forecasts of future demand 
in order to 
adjust their behavior accordingly.

\begin{figure}
    \centering
    \includegraphics[width=0.6\columnwidth]{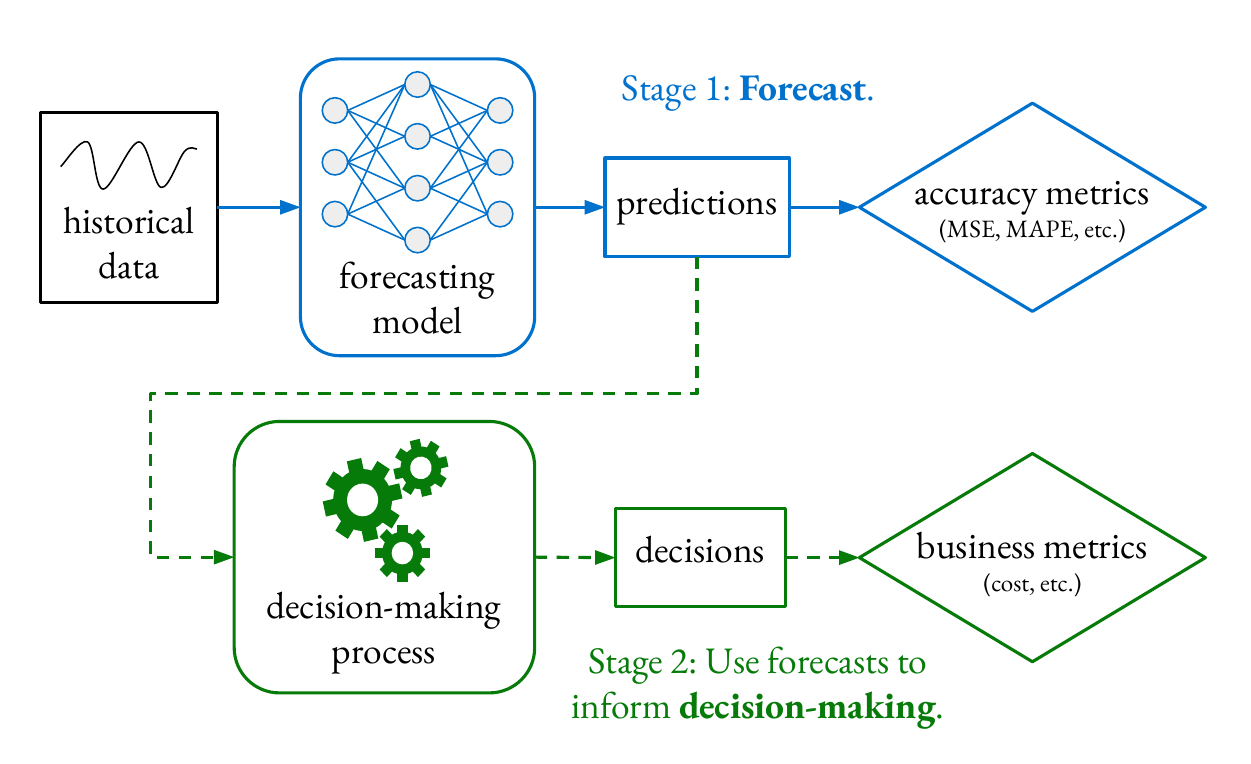}
    \caption{Typical separated estimation and optimization approach, where the learned forecaster (blue) is agnostic to business decision-making (green).
    Solid lines represent direct optimization. This work solidifies the dotted green lines.}
    \label{fig:flow_diagram}
\end{figure}

One common business problem reliant on time-series forecasts is
\emph{inventory management} \citep{chopra2007supply,syntetos2009forecasting}.
Here, businesses
must decide how much inventory to order
on a recurring basis,
balancing considerations such as
customer satisfaction,
costs 
of
holding surplus inventory (\emph{holding cost}), 
opportunity costs of out-of-stock items (\emph{stockout cost}), 
and keeping the supply chain running smoothly.
For example, grocery stores 
track their inventory,
anticipating demand and placing orders such that when customers demand e.g. toilet paper, they have enough stock to avoid lost sales and keep customers happy, while not having too much stock such that stale items are taking valuable shelf or warehouse space.
Since forecasts are used to decide how much inventory to order, the quality of forecasts can greatly influence downstream measures of 
business performance.
Typically, forecasters optimize and evaluate 
generic metrics agnostic to the downstream application, such as mean squared error (MSE) or mean absolute percentage error (MAPE). 
Assuming that these upstream forecasts are accurate, downstream decisions are subsequently treated as a separate step (Figure \ref{fig:flow_diagram}).

However, 
these 
generic metrics can be misaligned with
downstream business performance. 
For example, the 
business costs of over-forecasting and under-forecasting are often imbalanced (e.g. opportunity cost of lost sales could outweigh cost of holding extra inventory). Additionally, conventional forecasting metrics typically aim for a mean, median, or quantile of the distribution, without regard to the magnitude of fluctuations in predictions. Fluctuations in predictions can translate into fluctuations in orders, and as orders are passed upstream through the supply chain, uncertainties in forecasts can compound to create the bullwhip effect, an unstable and wildly oscillating demand
\citep{lee1997bullwhip,WANG2016691}.

One reason for the widespread
use of generic accuracy metrics such as MSE and MAPE is that downstream business metrics may be difficult to quantify or attribute to specific parts of the supply chain. Customer satisfaction, for example, might have a convoluted data generating process that is difficult to optimize directly.
However, as we show, optimization of generic 
metrics does not necessarily translate into improvements on downstream performance indicators. 

In this work, we propose a novel method for business metric-aware forecasting 
for inventory management systems. 
Our contributions include:
\begin{enumerate}
    \item Demonstrating that optimizing conventional metrics often translates into sub-optimal downstream performance.
    \item Deriving an efficient end-to-end differentiable procedure for optimizing forecasts for downstream inventory performance, compatible with any differentiable forecaster.
    \item Noting that downstream metrics are often at odds with one another, and proposing alternative combined objectives which trade off these metrics in different ways. 
    \item Empirically demonstrating the benefit of business metric-aware forecasting in univariate and multivariate datasets, under a variety of plausible downstream scenarios.
    \item Since time series datasets measuring demand are popular in the forecasting community, we release code\footnote{link excluded for anonymity} for others to evaluate the downstream utility of their forecasts.
\end{enumerate}

\section{Related Work}
\paragraph{Forecasting + Inventory Optimization}
Inventory management involves estimating future demand 
(\emph{forecasting}) and 
deciding how many orders to place 
to minimize costs and meet customer needs
(\emph{inventory optimization}).
Some works 
simply forecast the mean demand and treat it as the inventory optimization solution 
\citep{yu2013support,ali2013selecting}. 
However, this approach fails to account for imbalanced costs of over- and under-forecasting. 
Thus, in practice, it is common to take a \emph{separated estimation and optimization} approach \citep{turken2012multi,oroojlooyjadid2020applying}, which involves first (1) forecasting demand and then (2) plugging the estimates into inventory optimization (Figure \ref{fig:flow_diagram}).
However, 
in this approach, 
errors in forecasting 
and inventory optimization 
can compound.

One 
widely studied 
inventory model 
is the newsvendor problem, 
which involves determining the optimal order quantity for 
perishable or seasonal products 
to balance 
holding cost and
stockout cost.
In this setup, the optimal solution to the inventory optimization problem is a particular quantile of the demand distribution \citep{petruzzi1999pricing}. 
Thus, some works 
forecast various quantiles of the demand distribution \citep{10.14778/3137765.3137775,bertsimas2005data,taylor2000quantile}. Others 
use
feed-forward neural networks, kernel regression, and linear models to directly optimize these two costs \citep{ban2019big,oroojlooyjadid2017deep}.
We also optimize downstream inventory metrics directly, but 
allow more general cost objectives to be computed over the 
inventory system variables through use of differentiable simulation, making our approach applicable beyond newsvendor.

\paragraph{Demand Forecasting}
Demand forecasting is of interest in several businesses, 
including
retail 
\citep{fildes2022retail}, 
power grids 
\citep{ghalehkhondabi2017overview,suganthi2012energy}, 
emergency care \citep{jones2009multivariate}, and
municipal water \citep{donkor2014urban}.
Techniques for %
forecasting 
include
both classical statistical 
and
modern deep learning approaches.
Traditional time-series forecasting methods 
include autoregressive (AR) models \citep{box2015time,makridakis1997arma}, exponential smoothing \citep{gardner1985exponential,winters1960forecasting},
and the Theta model 
\citep{ASSIMAKOPOULOS2000521,hyndman2003unmasking}.
Deep learning architectures for time-series forecasting include convolutional neural networks \citep{bai2018empirical,oord2016wavenet}, recurrent neural networks \citep{hochreiter1997long,SALINAS20201181,rangapuram2018deep}, and attention-based methods \citep{fan2019multi,li2019enhancing,LIM20211748}, among others \citep{oreshkin2019n,challu2022n}. 
However, these works all 
optimize
business-agnostic metrics. 

\paragraph{Inventory Optimization} Given demand forecasts, the decision of how many orders to place could be treated as a constrained optimization problem \citep{dai2021neural}, a supervised deep learning problem 
\citep{qi2023practical}, or a reinforcement learning problem \citep{oroojlooyjadid2017deep}. There are also several common practices for placing orders \citep{inventorypolicies1968}.
For example, the (T, S) policy places orders every T days, and orders up to an inventory level S. 
\citet{PETROPOULOS2019251} explored the inventory performance of several
traditional forecasting models 
when a fixed periodic order-up-to (T, S) policy is used, finding that methods based on combinations had
superior inventory performance.
We use the same order-up-to policy in this work,
but instead of taking a separated estimation and optimization approach, we use differentiable simulation to optimize downstream business performance end-to-end.

\paragraph{Forecasting Competitions}
Forecasting competitions such as the M-Competitions \citep{MAKRIDAKIS2000451_M3,MAKRIDAKIS202054_M4,MAKRIDAKIS20221346_M5}
and the Favorita 
Competition \citep{favorita_data} have become popular benchmarks for 
development of modern time-series forecasting methods. While a substantial portion of this data is
industry time-series,
evaluation of model performance is largely conducted using
generic error metrics
which ignore downstream business performance. 
For example, the M3 competition 
measured performance using versions of
symmetric mean/ median absolute percentage error (sMAPE)
and
median relative absolute error. 
Submissions to the Favorita 
competition 
were evaluated using the normalized weighted root mean squared logarithmic error.
By releasing inventory performance code,
would like to 
further 
challenge researchers to make 
high-\emph{utility} 
predictions on this data.

\paragraph{Inventory Performance Metrics}
Across several inventory optimization applications, ranging from auto parts suppliers \citep{qi2023practical}, to online fashion retailers \citep{ferreira2016analytics}, to drug inventories \citep{dhond2000data}, 
the common objectives of interest are typically a 
function of stockout cost 
and holding cost.
These costs may be computed across 
various
lead times, %
different parts of the supply chain, or simply 
based on
historical data. High variance of orders has also been identified as an undesirable phenomenon due to the bullwhip effect in supply chains \citep{lee1997bullwhip,PETROPOULOS2019251}, in which 
fluctuations in downstream demand can cause exaggerated order swings upstream in the supply chain
that result in
customer-upsetting stockouts and wasteful excesses. 

\section{Inventory Management}
\label{sec:inventory_management}
\begin{figure}
    \centering
    \includegraphics[width=0.6\columnwidth]{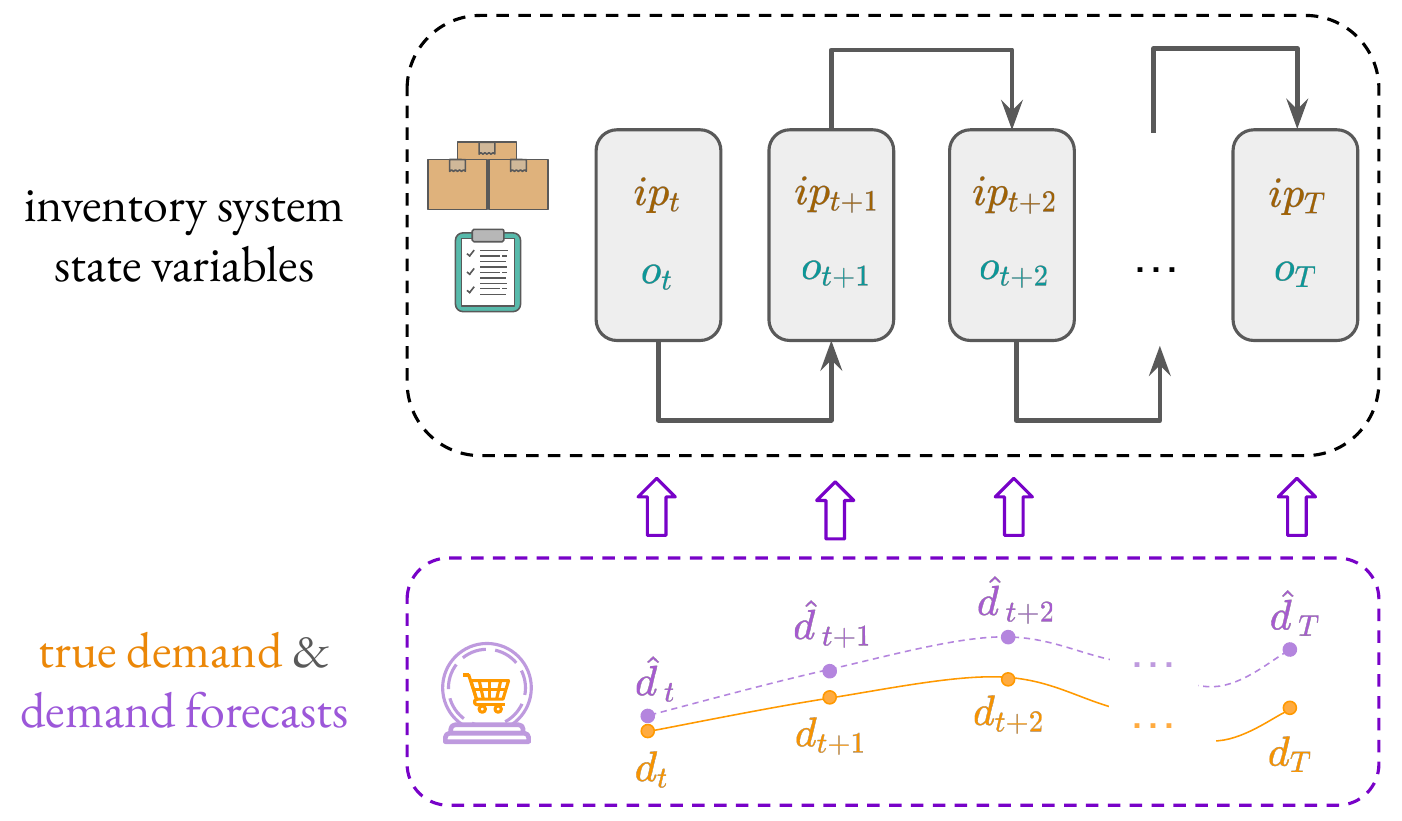}
    \caption{Inventory management systems keep track of state variables such as the current inventory position $ip$ and orders placed $o$. 
    Inventory position is decreased based on observed demand $d$, and replenished by orders $o$ which are placed based on demand forecasts $\widehat{d}$ and the current $ip$.
    }
    \label{fig:inventory_system}
\end{figure}

This section formalizes the quantities used and tracked in an inventory management system  (Figure \ref{fig:inventory_system}), and defines business-aware and business-agnostic metrics of interest.

\subsection{Formulation}\label{sec:system_definition}
For each time-series, we apply a rolling simulation approach 
in order to simulate an inventory system as it steps through each time point. 
Consider a time-series of the true demand $d_t$ at every time point $t = 1, 2, ..., T$. At each $t$, orders $o_t$ are placed with the expectation that they will take lead-time $L$ to come in. Using the order-up-to policy for inventory replenishment \citep{gilbert2005arima,PETROPOULOS2019251}, orders are given by: 
\begin{align}
    o_t = \widehat{D}_t^L + ss_t - ip_t \label{eq:orders}
\end{align}
where $\widehat{D}_t^L$ is the forecasted lead-time demand over the next $L$ timesteps, $ss_t$ is safety stock that adds a buffer to ensure that the orders placed cover the demand, and $ip_t$ is the inventory position. 
The \emph{true lead-time demand} $D_t^L$ and the \emph{forecasted lead-time demand} $\widehat{D}_t^L$ are given by:
\begin{align}
    D_t^L = \sum_{k=1}^L d_{t+k}, \hspace{3em} \widehat{D}_t^L = \sum_{k=1}^L \widehat{d}_{t, t+k} 
    \label{eq:demand}
\end{align}
where $\widehat{d}_{t, t+k}$ is the forecast of demand for time $t+k$ given data up to time $t$. 
\emph{Safety stock} is computed as follows:
\begin{align}
    ss_t = \Phi^{-1}(\alpha_s)\sigma_e,
    \label{eq:safety_stock}
\end{align}
where $\sigma_e$ is the standard deviation of the forecast errors, and $\Phi^{-1}(\alpha_s)$ is the inverse CDF of the normal distribution evaluated at some target service level $\alpha_s$. 
Assuming normally-distributed errors, with $\alpha_s$ probability, the safety stock plus lead time demand forecast should cover the actual demand.

\emph{Inventory position} $ip_t$ is obtained by taking the previous inventory position, adding the orders $o_{t-1}$ from the previous timestep, and subtracting the current demand $d_t$:
\begin{align}\label{eq:inventory_position}
    ip_t = ip_{t-1} + o_{t-1} - d_t.
\end{align}
We assume $ip_0=0$ and $o_0=0$.
Since orders take lead time $L$ to arrive, the inventory position can be decomposed into a sum of (a) how much inventory is actually on hand, termed \emph{net inventory level} $i_t$, and (b) how much inventory is on the way, termed \emph{work-in-progress level} $w_t$:
\begin{itemize}
    \item Inventory position: $ip_t = i_t + w_t$
    \item Net inventory: $i_t = i_{t - 1} + o_{t - L} - d_t$	
    \item Work-in-progress:	$w_t = w_{t-1} + o_{t-1} - o_{t-L}$
\end{itemize}

In summary, at each time $t$, orders $o_t$ are placed based on forecasted lead-time demand $\widehat{D}_t^L$ and the current inventory position $ip_t$. The orders and current demand then adjust the inventory position
$ip_{t+1}$,
and this process repeats for the entire length $T$ of the time-series.

\subsection{Evaluation Metrics}\label{sec:evaluation_metrics}

We evaluate and optimize both downstream business performance and conventional generic forecasting metrics.

\subsubsection{Downstream Inventory Performance}
One straightforward way to balance excess inventory, lost sales, and stability of orders is to frame everything in terms of cost. Thus we introduce a total cost metric, measured in units of money.
We also introduce a unitless metric, relative root-mean-square, which compares the performance versus a simple baseline. 

\emph{Total cost} (TC) is defined as a combination of the cost of holding excess inventory (holding cost $C_h$), the opportunity cost of running out of stock (stockout cost $C_s$), and the cost of fluctuations in the supply chain (order variance cost $C_v$):
\begin{align*}
    C_h &= c_h \cdot \mathbb{E}_{t}[\text{max}(0, i_t)]\\
    C_s &= c_s \cdot \mathbb{E}_{t}[\text{max}(0, -i_t)]\\
    C_v &= c_v \cdot \text{Var}_{t}(o_t)\\
    TC &= C_h + C_s + C_v,
\end{align*}
where expectations are taken over all time points $t$, and
$c_h, c_s, c_v \geq 0$ are constants. Specifically, $c_h$ is the unit holding cost, $c_s$ is the unit stockout cost, and $c_v$ is the unit order variance cost. If this information is available in a given problem setting, one can directly plug it in. Otherwise, practitioners can choose how to balance these different factors based on their domain expertise. For example, one might have the intuition that sales lost are more expensive per unit than the cost of holding an extra unit of inventory. If the different components of a supply chain are well-integrated so that the compounded uncertainty is not a major concern, the unit order variance cost may not need to be large. 

For settings in which the cost tradeoffs 
may be unknown,
we introduce the \emph{relative root-mean-square} (RRMS) metric:
\begin{align*}
    RRMS &= \sqrt{\text{rel}(C_h)^2 + \text{rel}(C_s)^2 + \text{rel}(C_v)^2},
\end{align*}
where relative performance to a naive baseline is defined as 
\begin{align*}
    \text{rel}(x) &= \sigma\left(\frac{x - x_{naive}}{x_{naive}}\right),
\end{align*}
where $\sigma$ is a sigmoid function.
The naive baseline we use in our experiments is a model which simply outputs the previous observation from one period ago. Note that the $rel(x)$ is a unitless quantity, as the unit costs cancel out in the numerator and denominator.

\subsubsection{Generic Forecasting Metrics}
We also evaluate and optimize
generic
forecasting metrics 
to understand the extent to which they
might indirectly optimize for downstream 
performance.
\emph{Mean squared error} (MSE) is given by averaging the squared error over time points 1 to $T$ and forecasting horizons 1 to $H$:
\begin{align*}
    \text{MSE} &= \frac{1}{TH} \sum_{t=1}^T \sum_{k=1}^H \left (d_{t+k} - \widehat{d}_{t, t+k} \right )^2.
\end{align*}

\emph{Symmetric mean absolute percentage error} (sMAPE) is: 
\begin{align*}
    \text{sMAPE} = \frac{1}{TH} \sum_{t=1}^T \sum_{k=1}^H \frac{|d_{t+k} - \widehat{d}_{t, t+k}|}{|d_{t+k}| + |\widehat{d}_{t, t+k}|} \times 2.
\end{align*}

\section{Methods}
Several challenges arise in 
optimization of
downstream business metrics. 
Here, we describe how persistent state variables can be computed
differentiably,
how to optimize for objectives that are computed over the entire time-series of these state variables rather than point-wise, how to provide additional supervision for univariate time series,
and how to simulate how models would be updated over time as new data points are observed. 
Additionally, we describe the 
models, %
datasets, 
and experiment setup. 

\subsection{Differentiable Computation of Metrics}\label{sec:diffable_comp}
For typical forecasting metrics (e.g. MSE, sMAPE, etc.), differentiable computation is relatively straightforward. 
These metrics can usually be decomposed such that at each time point, some differentiable quantity  (e.g. squared difference of prediction and actual value) is computed, and an average over time points is taken.
Computing inventory performance (e.g. total cost, RRMS), however, is less straightforward as there are persistent inventory state variables (e.g. net inventory level, orders) that must be tracked over time. Although inventory performance is not in general a differentiable quantity, we derive a series of computations which can simulate the inventory management system and order-up-to policy 
described by \eqref{eq:orders}--\eqref{eq:inventory_position}
in a differentiable manner. Composing this system with the outputs of a differentiable forecaster, we create an end-to-end differentiable system.

Naively, one could iterate over each time point, and apply the recursive inventory state update equations \eqref{eq:orders}--\eqref{eq:inventory_position}. 
However, for long time-series this would be computationally infeasible, due to 
the instability of backpropagation through a long chain of dependent states \citep{pascanu2013difficulty}.
Instead, assuming all quantities at time $t<0$ are 0, we show by expanding out the recursion (derivations in  Appendix) 
that the orders at any time $t$ can be written in closed form:
\begin{align}\label{eq:condensed_orders}
o_t = (\widehat{D}_t^L - \widehat{D}_{t-1}^L) + \Phi^{-1}(\alpha_s) \cdot (\sigma_{e,t} - \sigma_{e, t-1}) + d_t
\end{align}
and the net inventory at time $t$ can be written 
as:
\begin{align}
    i_t 
    &= \widehat{D}_{t-L}^L + \Phi^{-1}(\alpha_s) \cdot \sigma_{e,t-L} - \sum_{a=t-L+1}^t d_{a}. \label{eq:condensed_inventory_level}
\end{align}

These closed form equations are much more efficient to implement in terms of tensor operations than the original recursive equations,
and they allow us to simultaneously compute the net inventory levels and orders at all times given a tensor of demand forecasts at all times (detailed walkthrough in the Appendix). %

Given the inventory state variables $o_t$ and $i_t$ for all $t$, it is now feasible to compute metrics on top of these variables. Holding and stockout costs can be computed by applying a ReLU activation over $i_t$ and $-i_t$, and computing the average over timepoints. Variance of orders can be computed by averaging the squared difference between orders and the average number of orders. Finally, combinations or simple differentiable functions of these quantities can straightforwardly computed to yield both total cost (TC) and the relative root-mean-square metric (RRMS) (details in Appendix). 

\subsection{Double-Rollout Supervision}\label{sec:double_rollout}
Another challenge of optimizing downstream inventory performance is that
some 
aspects
must be 
computed 
holistically across multiple time points (e.g. order variance). 
For univariate time-series where a local model is trained on only one time-series, this is especially challenging due to limited supervision. To provide more supervision, we use a custom training method where at each time point, an inventory system simulation is rolled out over the next $H$ time points, where forecasting horizon $H > L$. 
By simulating the inventory system several times using different starting points in the univariate time-series, one can obtain several evaluations of TC and RRMS to serve as supervision from just one time-series (see Appendix for diagram).
For multivariate time-series, instead of forecasting for a horizon $H > L$ and then unrolling lead-time demands across that horizon, only forecasts of the requisite lead time $L$ are made since the other time-series can provide supervision, and double-rollouts are more computationally expensive.

\subsection{Roll-Forward Evaluation}\label{sec:roll_forward}
In real-world settings, as new data are 
collected, forecasting models are updated and decisions are made accordingly.
To simulate this process, 
we employ a 
training
procedure which rolls forward in time. For each time point from $t = 1$ to $t = T$, 
the model is trained with data up to $t$ 
using double-rollout supervision for univariate time-series,
and single-rollout supervision for multivariate time-series. 
Then, the model forecasts the next $L$ timesteps after $t$, i.e. $\widehat{d}_{t, t + k}$ for all $k \in \{1, 2, ..., L\}$. After all $T$ timesteps have been trained on and forecasted from, giving a $N \times T \times L$ tensor, inventory performance is computed over the $T$ timepoints. 
For each dataset we designate training, validation, and test time ranges, 
where 
validation 
data
is used for hyperparameter tuning, 
and 
test 
data
is used for reporting final performance.

\subsection{Models}\label{sec:models}
We explore two differentiable models for forecasting: (1) a seasonal scaling model, and (2) an LSTM encoder-decoder model. For univariate time-series, one local model is trained per time-series, and for multivariate time-series, one global model is trained across all time-series. Hyperparameter and model training details are in the Appendix.

\paragraph{Naive Seasonal Scaling Model}
This model has one learnable parameter $\beta \in \mathbb{R}$, the amount to scale observations from one period $P$ ago. That is, 
$\widehat{d}_{t, \text{seasonal scaler}} = \beta \cdot d_{t - P}.$
This model is valuable from an interpretability perspective, as $\beta > 1$ could indicate a preference towards over-forecasting versus the previous period of data, and $\beta < 1$ could indicate a preference towards under-forecasting. 

\paragraph{LSTM Encoder-Decoder}
This model 
has
an LSTM encoder which sequentially encodes a window of inputs, and an LSTM decoder which sequentially decodes to yield predictions across a forecasting horizon. For multivariate time-series, the covariates are embedded 
before being fed into the encoder, and a linear layer is used on top of the outputs of the decoder to yield the forecasts. See the Appendix for a diagram of the model architecture. 
\subsection{Data}\label{sec:data}

\subsubsection{M3 Monthly Industry Subset (Univariate)}
The monthly industry subset of the M3 competition data \cite{m3_data} consists of 334 univariate time-series with up to 144 time points, where time points occur on a monthly basis. As described by \citet{PETROPOULOS2019251}, this subset can serve as a proxy for demand on a monthly basis. These time-series are not aligned 
by start date,
have varying lengths, and are not directly related to each other. Hence, each time-series in this dataset is treated separately as a univariate time-series for modeling purposes.

\subsubsection{Favorita Grocery Sales (Multivariate)} 
The Corporaci\'{o}n Favorita Grocery Sales Forecasting dataset \cite{favorita_data} consists of sales data across several stores and products. The dataset includes covariates such as oil prices, location, day of week, month, and holidays. We use a similar preprocessing pipeline as in \citet{LIM20211748} to yield 90,193 distinct time-series with up to 396 time points, where time points occur on a daily basis from 2015 to 2016. As grocery replenishment often occurs on a daily basis, the inventory system is updated daily. These time-series are aligned to start at the same time in the real world, missing values are imputed with zeros, and the time-series are likely correlated with each other since they are all associated with sales in Corporaci\'{o}n Favorita. Thus, this dataset is treated as a multivariate time-series dataset,
and one global model is learned.

\subsection{Experiment Setup}\label{sec:experiment_setup}
The models are optimized using the mean squared error (MSE), relative root mean square (RRMS), and total cost (TC) objectives across several settings of unit costs.

For M3, a separate local model is trained with double-rollout supervision and roll-forward evaluation for each of the 334 univariate time-series. Since each time point corresponds to one month, a periodicity of $P = 12$ is used for seasonal models. An encoding window of 24 months is used as input to the model, allowing the model to learn use the previous two periods of history for its predictions. Predictions are made for a forecasting horizon of 12 months, so that the double-rollout can compute inventory performance over multiple time points. A lead time of $L=6$ months is used. Out of 144 months, forecasting models are initially trained with 72 months, then validated until 108 months, and then tested until 144 months. 
Since the safety stock discouraged forecasting errors, whereas a high or low unit holding cost could encourage over- or under-forecasting, we choose to have $\alpha_s = 0.5$, so $ss_t = 0$ for all $t$, to avoid unstable interactions between unit costs and safety stocks. For the TC objective, models are trained on every combination of unit holding costs $c_h \in \{1, 2, 10\}$, unit stockout costs $c_s \in \{1, 2, 10\}$, and unit order variance costs 
$c_v \in \{1\mathrm{e}{-6}, 1\mathrm{e}{-5}\}$ (chosen based on the order of magnitude of demand).

For Favorita, one global model is trained with single-rollout supervision and roll-forward evaluation for all 90,193 multivariate time-series. A global model is used because all time-series are aligned and correlated with one another, and training a separate model for each series would be computationally expensive. Each time point corresponds to one day, so a periodicity of $P = 7$ is used. 
An encoding window of 90 days is input to the model, which forecasts the next 30 days. A lead time of $L=7$ days is used. Out of 396 days, the training cutoff is at day 334, the validation cutoff is day 364, and the remainder is used for testing. Again, we set $\alpha_s = 0.5$. For the TC objective, due to more expensive training, a subset of $N=10,000$ samples are extracted to test all combinations of unit holding costs $c_h \in \{1, 2, 10\}$, unit stockout costs $c_s \in \{1, 2, 10\}$, and unit order variance costs $c_v \in \{1\mathrm{e}{-3}, 1\mathrm{e}{-2}\}$.

\section{Results}

\paragraph{Unit Cost-Agnostic Performance} Tables \ref{tab:m3_mse} and \ref{tab:favorita_mse} characterize the performance of several forecasters trained on the full M3 and Favorita datasets when evaluated on typical forecasting metrics, MSE and sMAPE, and an inventory performance metric, RRMS.
All of the models in these tables are trained and evaluated on objectives that are agnostic to unit costs $c_h, c_s,$ and $c_v$.

In both M3 and Favorita, the Seasonal Scaler and LSTM models trained with MSE objective perform competitively with classical models on MSE and sMAPE, either performing better than or within the range of performance spanned by the ARIMA, Exponential Smoothing, and Theta models.
In the M3 dataset, the model with best RRMS is the seasonal scaler trained on RRMS---abbreviated as Seasonal Scaler (RRMS). However, it achieves worse MSE (22.10$\times 10^5$) than the Seasonal Scaler (MSE), LSTM (MSE), Exponential Smoothing, and Theta models (MSEs ranging 13.82$\times 10^5$ to 20.89$\times 10^5$). In the Favorita dataset, the Seasonal Scaler (MSE) and the Seasonal Scaler (RRMS) outperform all other models on RRMS, despite having worse MSE (1.23$\times 10^2$ and 1.25$\times 10^2$) than the LSTM (MSE), ARIMA, and Theta models (0.88$\times 10^2$ to 1.12$\times 10^2$). On the M3 dataset, the LSTM (MSE) achieves the best MSE (13.82$\times 10^5$), yet has the second to worst RRMS (1.23). On Favorita, the LSTM (MSE) objective again achieves the best MSE (0.88$\times 10^2$), yet has the worst RRMS (1.17). Overall, performance on typical forecasting metrics (MSE and sMAPE) appears misaligned with relative inventory performance (RRMS), and optimizing for one does not inherently seem to optimize for the other.

\begin{table}
    \centering
    \caption{M3 test performance of models that are agnostic to unit costs. Note that the best-performing models on MSE are misaligned with the best-performing models on RRMS.}
    \small
    \begin{tabular}{lccc}
\toprule
     Model (Objective) &  MSE &  sMAPE &  RRMS \\
     & ($\times 10^{-5}$) & &\\
\midrule
            Seasonal Scaler (MSE) &                   20.89 &   0.27 &  1.27 \\
            Seasonal Scaler (RRMS) &                   22.10 &   0.28 &  \textbf{0.74} \\
            LSTM (MSE) &                   \textbf{13.82} &   0.24 &  1.23 \\
          LSTM (RRMS) &                  226.49 &   1.25 &  1.02 \\
                 ARIMA &                   25.78 &   0.34 &  1.12 \\
             Exponential Smoothing &                   15.21 &   0.24 &  1.12 \\
                 Theta &                   14.00 &   \textbf{0.22} &  1.09 \\
\bottomrule
\end{tabular}

    \label{tab:m3_mse}
\end{table}

\begin{table}
    \centering
    \caption{Favorita test performance of models that are agnostic to unit costs. Note the best-performing models on MSE are misaligned with the best-performing models on RRMS.}
    \small
    \begin{tabular}{lccc}
\toprule
     Model (Objective) &  MSE &  sMAPE &  RRMS \\
     & ($\times 10^{-2}$) & & \\
\midrule
 Seasonal Scaler (MSE) &                    1.23 &   \textbf{1.51} &  \textbf{0.85} \\
 Seasonal Scaler (RRMS) &                    1.25 &   1.73 &  0.94 \\
            LSTM (MSE) &                    \textbf{0.88} &   1.77 &  1.17 \\
           LSTM (RRMS) &                    3.42 &   2.84 &  1.10 \\
 ARIMA &                    1.12 &   1.76 &  1.06 \\
 Exponential Smoothing &                    1.36 &   1.80 &  1.10 \\
                 Theta &                    1.08 &   1.81 &  1.09 \\
\bottomrule
\end{tabular}

    \label{tab:favorita_mse}
\end{table}

\begin{table}
    \centering
    \caption{M3 test total cost across several unit cost settings $(c_h, c_s, c_v)$.}
    \small

\begin{tabular}{lcccccc}
\toprule
     Model (Objective) & (1, 1, 1e-05) & (1, 1, 1e-06) & (1, 10, 1e-05) & (1, 10, 1e-06) & (10, 1, 1e-05) & (10, 1, 1e-06) \\
\midrule
            LSTM (MSE) &          5,826 &          5,435 &          39,519 &          39,128 &          20,660 &          20,269 \\
           LSTM (RRMS) &         17,188 &         17,102 &         170,738 &         170,652 &          17,474 &          17,388 \\
             LSTM (TC) &          6,390 &          5,997 &          36,086 &          35,805 & \textbf{10,146} & \textbf{10,400} \\
 Seasonal Scaler (MSE) &          5,680 &          5,372 &          43,791 &          43,483 &          15,610 &          15,302 \\
Seasonal Scaler (RRMS) &          5,771 &          5,476 &          45,216 &          44,920 &          15,314 &          15,018 \\
  Seasonal Scaler (TC) & \textbf{5,185} & \textbf{4,884} & \textbf{35,268} & \textbf{34,918} &          12,178 &          11,996 \\
\bottomrule
\end{tabular}

    \label{tab:m3_tradeoffs}
\end{table}

\begin{table}
    \centering
    \caption{Favorita test total cost across several unit cost settings $(c_h, c_s, c_v)$.}
    \small

\begin{tabular}{lcccccc}
\toprule
     Model (Objective) &  (1, 1, 1e-02) &  (1, 1, 1e-03) &  (1, 10, 1e-02) &  (1, 10, 1e-03) & (10, 1, 1e-02) & (10, 1, 1e-03) \\
\midrule
            LSTM (MSE) &          26.94 &          20.30 &          158.65 &          152.01 &          71.25 &          64.61 \\
           LSTM (RRMS) &          94.51 &          72.08 &          162.55 &          140.12 &         652.76 &         630.33 \\
             LSTM (TC) &          24.69 & \textbf{18.39} & \textbf{115.72} &          116.40 &          35.96 & \textbf{29.72} \\
 Seasonal Scaler (MSE) &          24.52 &          21.22 &          200.51 &          197.22 &          36.25 &          32.96 \\
Seasonal Scaler (RRMS) & \textbf{23.40} &          18.69 &          158.51 &          153.80 &          51.84 &          47.13 \\
  Seasonal Scaler (TC) &          23.52 &          18.51 &          118.48 & \textbf{107.03} & \textbf{32.68} &          30.18 \\
\bottomrule
\end{tabular}

    \label{tab:favorita_tradeoffs}
\end{table}

\begin{figure}
\centering
    \begin{subfigure}[b]{0.342\columnwidth}
         \centering
         \includegraphics[width=\columnwidth]{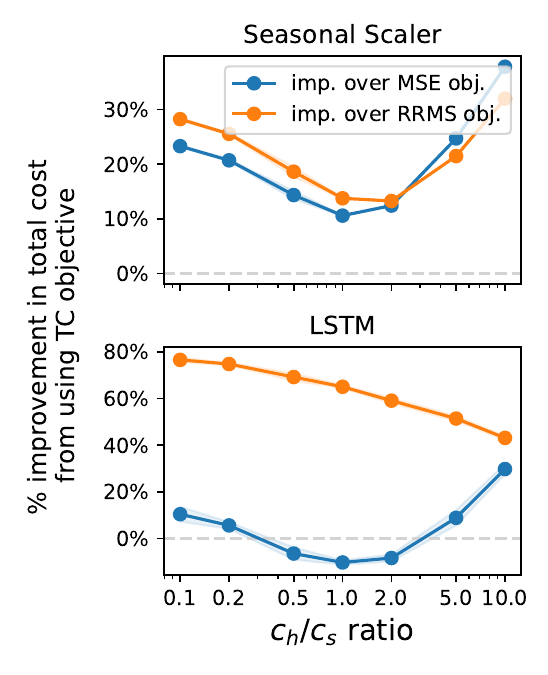}
        \caption{M3}
        \label{fig:m3_improvement}
    \end{subfigure}    
    \begin{subfigure}[b]{0.375\columnwidth}
         \centering
         \includegraphics[width=\columnwidth]{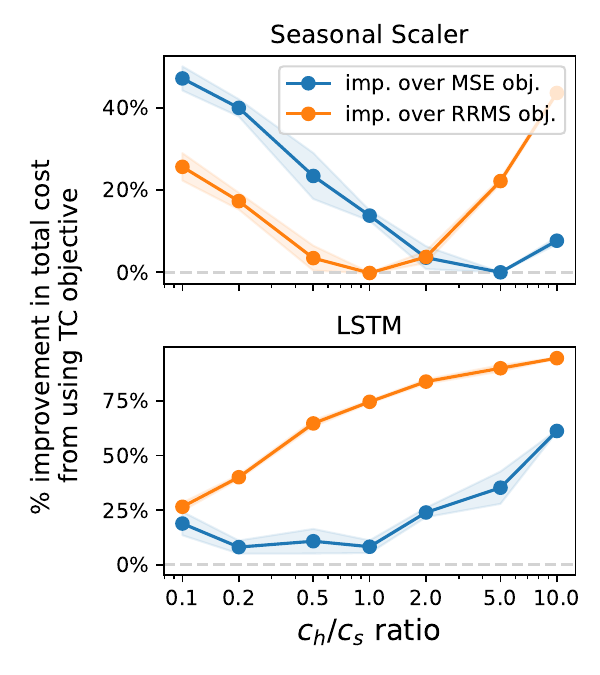}
        \caption{Favorita}
        \label{fig:favorita_improvement}
    \end{subfigure}
    \caption{
    Average relative percentage improvement in test total cost from using the TC objective over the MSE objective and RRMS objective across various $c_h/c_s$ ratios.
    95\% CI are computed across different $c_v$ values.}
    \label{fig:improvement}
\end{figure}

\begin{figure}
\centering
    \begin{subfigure}[b]{0.35\columnwidth}
         \centering
         \includegraphics[width=\textwidth]{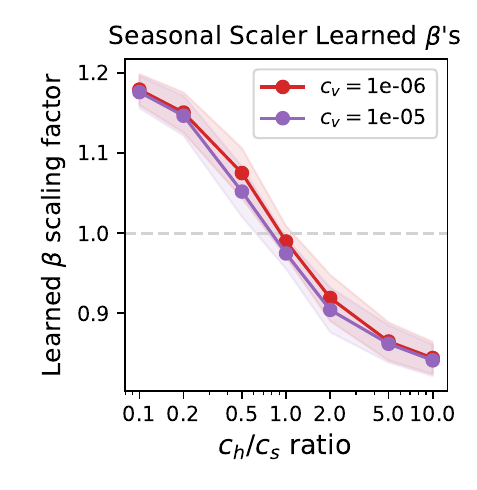}
        \caption{M3}
        \label{fig:m3_betas}
    \end{subfigure}    
    \begin{subfigure}[b]{0.35\columnwidth}
         \centering
         \includegraphics[width=\textwidth]{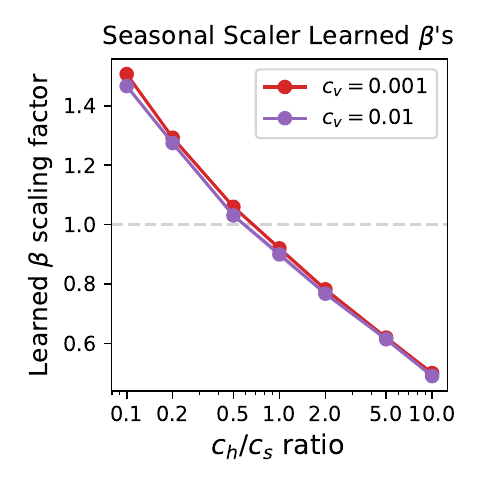}
        \caption{Favorita}
        \label{fig:favorita_betas}
    \end{subfigure}
    \caption{Learned scaling factors for the naive seasonal scaler on the M3 (left) and Favorita (right) datasets, across several unit cost tradeoffs. Dotted line corresponds to $\beta = 1$.}
    \label{fig:betas}
\end{figure}

\paragraph{Performance Across Several Unit Cost Tradeoffs} 

Tables \ref{tab:m3_tradeoffs} and \ref{tab:favorita_tradeoffs} contain the test total cost across various unit cost settings. While Seasonal Scaler observes some benefit from training using the RRMS objective, it appears to cause unstable performance for the LSTM. 
On the other hand, 
\emph{using the TC objective almost always improves the total cost} of the Seasonal Scaler and LSTM models, except for the LSTM on the M3 dataset, where a more consistent benefit is observed for imbalanced $c_h$ and $c_s$ (Figure \ref{fig:improvement} and Appendix Figure \ref{fig:both_circles}). The greater the imbalance in $c_h$ and $c_s$, the greater the improvement from using TC objective. For example, on Favorita, the Seasonal Scaler trained on TC achieves a 45.7\% improvement over that trained by MSE when $(c_h, c_s, c_v) = (1, 10, 10^{-3})$, and the LSTM encoder-decoder trained on TC achieves a 54.0\% improvement over that trained by MSE when $(c_h, c_s, c_v) = (10, 1, 10^{-3})$.

Leveraging the interpretability of the Seasonal Scaler model, we graph the relationship between the learned $\beta$s and the tradeoffs between $c_h$ and $c_s$ (Figure \ref{fig:betas}). In both M3 and Favorita, with larger $c_h/c_s$ ratios and increasing $c_v$, the learned $\beta$ scaling factor decreases. 

Another benefit of end-to-end optimization of forecasts is interpretability that the forecasts themselves provide. Appendix Figure \ref{fig:avg_demand} plots the forecasted and true lead demands, averaged over all series in each dataset, for various cost objectives. When trained on MSE, the LSTM model tends to slightly over-forecast the true demand in aggregate, whereas the Seasonal Scaler model appears to match or slightly under-forecast. When trained on the TC objective, both models tend to under-forecast when the $c_h/c_s$ ratio is high,
and over-forecast when the ratio is low.
The LSTM predictions on M3 are smoother than that of the Seasonal Scaler, perhaps because the LSTM is more flexible and able to reduce the variance of its predictions in order to help reduce order variance, whereas the Seasonal Scaler can only scale the previous period by some constant (but note that due to roll-forward evaluation, where the model is updated each time point, this constant can change over time). The LSTM predictions on Favorita are more variable, perhaps due to the small order variance penalty.

\section{Discussion}
We demonstrate the limitations of using standard forecasting metrics that are agnostic to downstream business metrics, and propose a method for augmenting models with business metric-aware objectives. 
Common forecasting metrics such as MSE and sMAPE can be misaligned with downstream inventory performance, and optimizing for such metrics does not inherently optimize for inventory performance metrics (Tables \ref{tab:m3_mse} and \ref{tab:favorita_mse}).
We derive a differentiable procedure for computing inventory performance, and demonstrate that especially when costs are imbalanced, utilizing a business metric-aware total cost objective often significantly improves downstream costs (Tables \ref{tab:m3_tradeoffs} and \ref{tab:favorita_tradeoffs}, Figures \ref{fig:improvement} and \ref{fig:both_circles}). 

When deployed in a roll-forward evaluation framework, we observe that the Seasonal Scaler can be surprisingly effective (Table \ref{tab:m3_mse} and \ref{tab:favorita_mse}) in some cases despite only having one learned parameter. One possible explanation is that the learned constant can vary over time and adapt to new data each timestep as it is observed. In contrast to standard evaluation in which one assumes that the model is learned on data from a fixed time period and evaluated on a fixed test time period, this form of evaluation could be more realistic for the inventory management setting, where forecasts are constantly updated to inform daily, weekly, or monthly decisions. At the same time, the Seasonal Scaler must have an output proportional to the previous period which restricts the flexibility of this model class even if the proportion can change over time. The more flexible LSTM model performs best in MSE and in several of the cost tradeoff settings.

There are also some practical benefits of end-to-end optimization with business metric-aware objectives. When demand forecasting and inventory optimization are treated separately (as is typical), errors in each component are likely to compound. 
While some have proposed
searching over conventional forecasting methods for models which happen to achieve better downstream inventory performance \citep{PETROPOULOS2019251}, this is computationally expensive as several models must be trained, each of which have their own hyperparameters to be tuned. By directly optimizing end-to-end, this can save computation. Additionally, 
our methods for optimizing 
inventory performance 
are compatible with any differentiable forecaster.

More broadly, 
business metric-aware forecasting
could
be useful beyond 
inventory management, for other
business problems 
that rely on forecasts. 
Through our case study in inventory, we have shown how to tackle some common challenges that might arise from attempting to simulate downstream decision-making processes and systems, including: 
(1) differentiable computation of persistent state variables,
(2) optimization of objectives that must be computed over the entire time-series rather than point-wise, 
(3) providing additional supervision with limited time series,
and (4) simulating how models would be updated over time as new data points are observed. 
Business decisions that rely on forecasts may inherently prefer 
error distributions that are biased in certain ways, 
and we encourage others to explore business metric-aware forecasting in their own business problems.

\paragraph{Limitations and Future Work}
While TC outperforms MSE when costs are imbalanced, when costs balance each other out, the MSE objective can perform comparably or even slightly better than the TC objective
(Figure \ref{fig:m3_improvement}, bottom). 
The TC objective, 
while differentiable, is more complex than MSE, and can be sensitive to hyperparameter tuning. 
Similarly, while RRMS is a convenient unitless objective, it can also be difficult to optimize.
While in this work we decided to purely compare inventory vs. generic objectives, future work might explore pre-training with MSE and fine-tuning with TC or RRMS. 

Finally, there are several possible avenues for further exploration.
Future work could use the lens of business metric-aware forecasting to consider other differentiable model architectures, time series datasets, downstream objectives, and downstream business problems.

\section{Acknowledgements}
We gratefully acknowledge Kin Olivares, the Google Cloud AI Discovery, and Google Cloud Research teams for helpful conversations and feedback. This material is based upon work supported by the National Science Foundation Graduate
Research Fellowship Program under Grant No. DGE1745016 and DGE2140739. Any opinions, findings,
and conclusions or recommendations expressed in this material are those of the author(s)
and do not necessarily reflect the views of the National Science Foundation.

\newpage
\bibliographystyle{apalike}
\bibliography{refs}

\newpage
\appendix
\section{Differentiable Computation of Inventory Performance Metrics}\label{app:differentiable}
In this appendix section we derive equations \eqref{eq:condensed_orders} and \eqref{eq:inventory_position}, and walk through differentiable computation of all inventory system variables.

Consider a model trained on $N$ univariate time-series, each with at most $T$ time points. Each time point, the model makes a lead-time forecast for the demand across a forecast horizon $L$. Thus, the model outputs a tensor $\widehat{d} = \mathbb{R}^{N \times T \times L}$, where an entry $\widehat{d}[i, t, l]$ corresponds to the forecasted demand in the $i$th series for time $t + l$ at time $t$.

\subsection{Computing forecasted lead-time demand}
The forecasted lead-time demand 
is
$\widehat{D}_t^L = \sum_{l=1}^{L} \hat{d}[:, t, l]$, i.e. summation along the last axis of the $\widehat{d}$ tensor.

\subsection{Computing orders}
Here we derive equation \eqref{eq:condensed_orders}.
Alternating plugging in the inventory position equation \eqref{eq:inventory_position} into the order-up-to policy equation \eqref{eq:orders} and recursively plugging in \eqref{eq:orders} to itself, we can expand the expression for orders into a closed form:
\begin{align*}
    o_t &= \widehat{D}_t^L + ss_t - ip_t \\
    &= \widehat{D}_t^L + ss_t - (ip_{t-1} + o_{t-1} - d_t) \\
    &= \widehat{D}_t^L + ss_t - ip_{t-1} - (\widehat{D}_{t-1}^L + ss_{t-1} - ip_{t-1}) + d_t \\
    &= (\widehat{D}_t^L - \widehat{D}_{t-1}^L) + (ss_t - ss_{t-1}) + d_t \\
    &= (\widehat{D}_t^L - \widehat{D}_{t-1}^L) + \Phi^{-1}(\alpha_s) \cdot (\sigma_{e,t} - \sigma_{e, t-1}) + d_t,
\end{align*}
where the last step plugs in the safety stock definition \eqref{eq:safety_stock}. Using this equation we have derived, it is now possible to compute $o_t$ given just a tensor of demand forecasts $\widehat{d}$ and true demands $d$.

\subsection{Computing net inventory}

Here we derive the closed-form net inventory equation \eqref{eq:condensed_inventory_level}. Recursively plugging in the net inventory equation as well as the closed form expression for orders \eqref{eq:condensed_orders}, we have:
\small
\begin{align*}
    i_t &= i_{t-1} + o_{t-L} - d_t\\
    &= (i_{t-2} + o_{t-1-L} - d_{t-1}) + o_{t-L} - d_t\\
    &= i_{0} + \sum_{j=0}^{t-L} o_{j} - \sum_{k=1}^{t} d_{k}\\
    &= i_{0} + \sum_{j=0}^{t-L} \left( (\widehat{D}_j^L - \widehat{D}_{j-1}^L) + (ss_{j} - ss_{j-1}) + d_j \right) - \sum_{k=1}^{t} d_{k}\\
    &= i_{0} + \sum_{j=0}^{t-L} (\widehat{D}_j^L - \widehat{D}_{j-1}^L + ss_{j} - ss_{j-1}) + d_0 - \sum_{k=t-L + 1}^{t} d_{k}\\
    &= i_{0} + \widehat{D}_{t-L}^L + \Phi^{-1}(\alpha_s) \cdot \sigma_{e,t-L} + d_0 - \sum_{k=t-L + 1}^{t} d_{k},
\end{align*}
\normalsize
assuming that all quantities at time $t=-1$ are equal to 0. Then, $i_0 = i_{-1} + o_{-L} - d_0 = -d_0$. Simplifying,
$$i_t = \widehat{D}_{t-L}^L + \Phi^{-1}(\alpha_s) \cdot \sigma_{e,t-L} - \sum_{k=t-L + 1}^{t} d_{k}.$$
Intuitively this makes sense, as the 
net inventory is determined by the lead time forecast from $L$ time steps prior, with additional safety stock estimated at the time, subtracting the interim demand leading up to the current time step.
This closed form equation is much more efficient to implement in terms of tensor operations than the original recursive equation for inventory position. We have already described how to compute the first term, and the last term is simply the sum of true demands in a window of size $L$ leading up to time $t$. The 2nd term (safety stock) is computed by some constant $\Phi^{-1}(\alpha_s)$ dependent on desired service level $\alpha_s$, multiplied by the standard deviation of forecast errors up until the previous time $\sigma_{e,t-1}$. 
Similar to the net inventory, we can also derive a closed-form expression for the inventory position:
$ip_t = \widehat{D}_{t-1}^L + \Phi^{-1}(\alpha_s) \cdot \sigma_{e,t-1} - d_{t}$,
and the work in progress can simply be derived as $w_t = ip_t - i_t$.

\subsection{Computing safety stock} 
The inverse CDF of the target service level $\Phi^{-1}(\alpha_s)$ is a constant and straightforward to compute. The standard deviations of forecast errors $\sigma_{e,t}$ are more involved since forecasts are made at each time point for some horizon, but can be computed as follows:
\begin{itemize}
    \item Create an $N \times T$ tensor $M$ where $M[:, t] = t$.
    \item Construct sliding windows of size $L$ along the time dimension. This will create a tensor $M'$ with the same shape as $\widehat{d}$ ($N \times T \times L$), of the time of each entry.
    \item Repeat the tensor $T$ times in a new (fourth) dimension, thresholding to create a binary mask $M''$ for each $t \in \{1, 2, ..., T\}$ corresponding to whether that time has occurred. That is, $M''[i,t,l,t'] = \mathbf{1}\left\{M'[i, t, l, t'] \leq t'\right\}$.  
    \item Compute per-element squared error $E$ of the $N \times T \times L$ predictions. Copy $T$ times to get $E'$ ($N \times T \times L \times T$).
    \item Multiply the repeated error $E'$ element-wise with the binary mask $M''$ (both should be $N \times T \times L \times T$). Sum along the second and third dimensions (of size $T$ and $L$), and divide by the sum of the binary mask along the second and third dimensions. This gives the average squared forecast errors for each time point, for each time-series. Take the square root to get the standard deviation.
\end{itemize}

\subsection{Computing inventory performance metrics}
There are three main aspects of inventory performance we examine:
\begin{enumerate}
    \item Holding cost: $C_h = c_h \cdot \mathbb{E}[\text{max}(0, i_t)]$
    \item Stockout cost: $C_s = c_s \cdot \mathbb{E}[\text{max}(0, -i_t)]$
    \item Order variance cost: $C_v = c_v \cdot \text{Var}(o_t)$
\end{enumerate}
The holding cost can be computed by passing the computed inventory positions through a ReLU activation, and then taking the average across times and series. The variance of orders can be computed by taking the average orders for each series, subtracting them from the orders, squaring, and then taking the expectation. 

The total cost (TC) and relative RMS (RRMS) objectives combine these three components in differentiable ways (either summation or subtracting and dividing by a constant, squaring, and summing). 

\newpage
\section{Double-Rollout Supervision}
\begin{figure}[h]
    \centering
    \includegraphics[width=0.5\columnwidth]{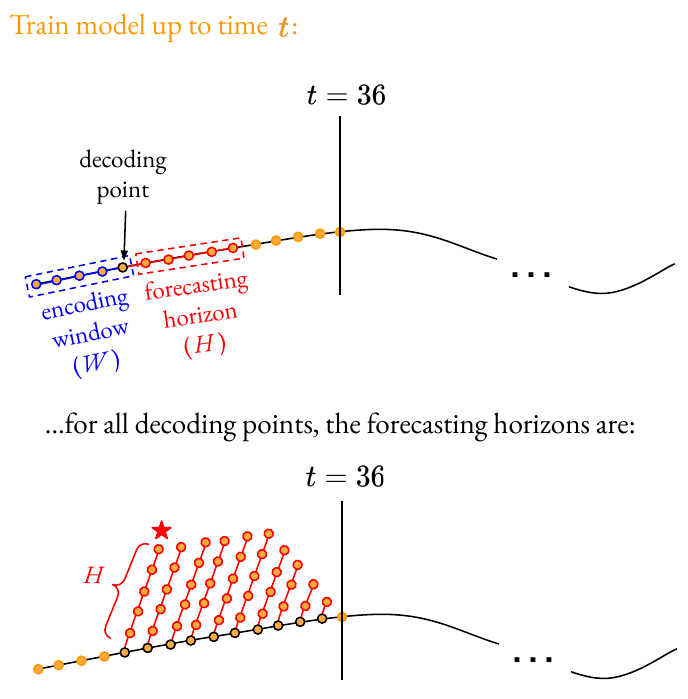}
    \caption{First step of the double-rollout training procedure. At each decoding point, the model forecasts a long forecasting horizon $H$.}
    \label{fig:double_rollout1}
\end{figure}

\begin{figure}[h]
    \centering
    \includegraphics[width=0.5\columnwidth]{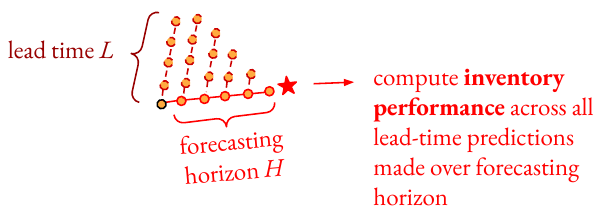}
    \caption{Second step of the double-rollout training procedure. For each forecasted horizon of $H$ time points, take a sliding window of size $L$ as the lead-time demand forecasts and compute inventory performance across the $H$ time points.}
    \label{fig:double_rollout2}
\end{figure}

Here we describe the double-rollout supervision technique. Since some objectives must be computed holistically across multiple time points (e.g. order variance), a single series of points may only yield one inventory performance value. If one model is being trained per time series (as is the case in M3 univariate data), this provides very little supervision for the model. The double-rollout supervision technique addresses this problem by having the model predict a long forecasting horizon $H$, which is then treated as the series to compute inventory performance over (Figure \ref{fig:double_rollout1}, top). A sliding window of size $L$ is taken over the $H$ time points  (Figure \ref{fig:double_rollout1}, bottom) in order to compute lead-time demands across the $H$ time points (Figure \ref{fig:double_rollout2}). Thus, we choose an $H > L$. If there are $t - W$ decoding points, then this gives $t-W$ series of length $\leq H$, which can provide $t-W$ measures of inventory performance to help supervise learning.

\newpage
\section{Roll-Forward Evaluation}
In real-world settings, models are updated over time as new data are collected. 
To better simulate the process of how forecasting models could feasibly be updated over time, we employ a training procedure which rolls forward in time. 

\begin{figure}[h]
    \centering
    \includegraphics[width=0.4\columnwidth]{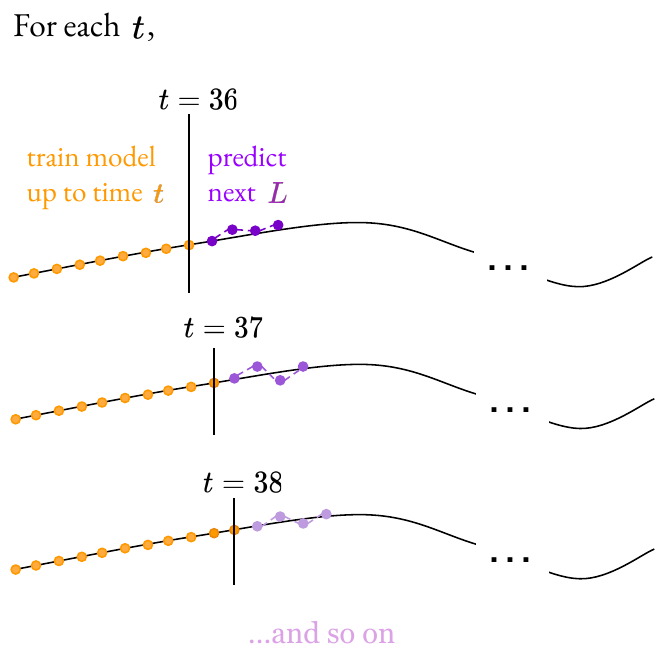}
    \caption{Each time $t$, the model is updated using all data available up until $t$, and forecasts the next $L$ time steps.}
    \label{fig:roll_forward1}
\end{figure}

\begin{figure}[h]
    \centering
    \includegraphics[width=0.5\columnwidth]{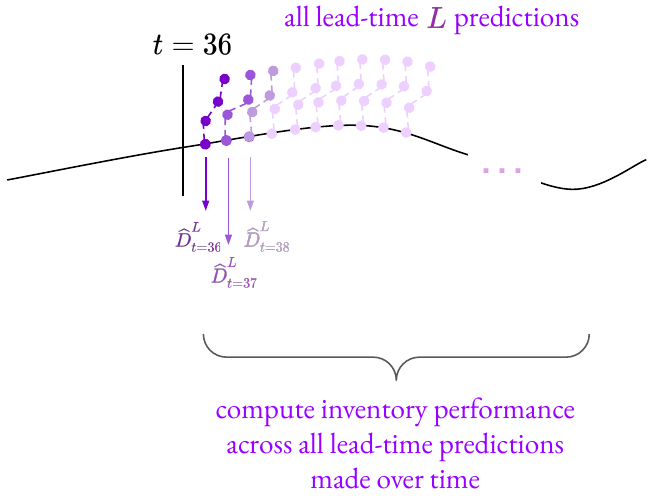}
    \caption{Predictions resulting from the roll-forward evaluation procedure after rolling across the whole series.}
    \label{fig:roll_forward2}
\end{figure}

At each time point $t = 1, 2, ..., T$, the model is updated with the most recent data by taking additional gradient steps based on all data up to time $t$ (i.e. fine-tuning to the latest dataset each time). Given a model trained on the most recent demand data up time $t$, predictions are made for the next lead-time $L$ time steps (Figure \ref{fig:roll_forward1}), giving $\widehat{D}^L_{t}$ (Figure \ref{fig:roll_forward2}).

These predictions are used by the order-up-to policy \eqref{eq:orders} in order to determine how many orders to place at that time.
This way, when inventory performance is computed across all $t$, the predictions are made using the most up-to-date models that could have been trained at each time point.

\newpage
\section{Hyperparameter Selection}

Ten iterations of random search with the MSE, TC, and RRMS objectives are used to choose hyperparameters for the M3 and Favorita datasets. For the Favorita dataset, a subset of 10,000 time series is used in order to tune hyperparameters more quickly. For the Naive Seasonal Scaler, the grid of hyperparameters considered is batchsize (100, 200, 300) and learning rate (0.01, 0.001, 0.0001). For the LSTM, the grid of hyperparameters considered is batchsize (100, 200, 300, 500), hidden size (32, 64, 128), and learning rate (0.01, 0.001, 0.0001, 0.00001). There is a small amount of subsequent manual tuning to try values beyond the grid if the best selected value was at the edge of the grid. 

For the M3 dataset, a hidden size of 20 is used for both the encoder and decoder. For the Favorita dataset, a hidden size of 64 is used for both the encoder and decoder, and the embedding size is 10. Categorical variables are embedded such that each possible value is stored as a different learnable 10-dimensional vector, and numerical variables are passed through a linear layer with 10-dimensional output. The \verb|run_all.sh| script in the included code supplement contains commands to run all of the experiments, with exact hyperparameter settings included.

\section{Training Details}
On Favorita, LSTM encoder-decoder models are trained on machines with A100 GPUs, and require about 10GB of memory. On M3, LSTM encoder-decoder models are trained on machines with 32 CPUs, and since one model is trained for each series, model training is done in parallel, taking about 2GB of memory at a time. All model training is done on the Ubuntu operating system. Package versions are included in the code package.

For M3, 
the training and validation cutoff points are chosen so that there are substantial time points in the training set (72) while still leaving enough to evaluate validation (36) and test (36) performance.
For Favorita, 
training and validation cutoffs are the same as in \citet{LIM20211748}.

\section{Architecture}

\begin{figure}[h]
    \centering
    \includegraphics[width=0.6\columnwidth]{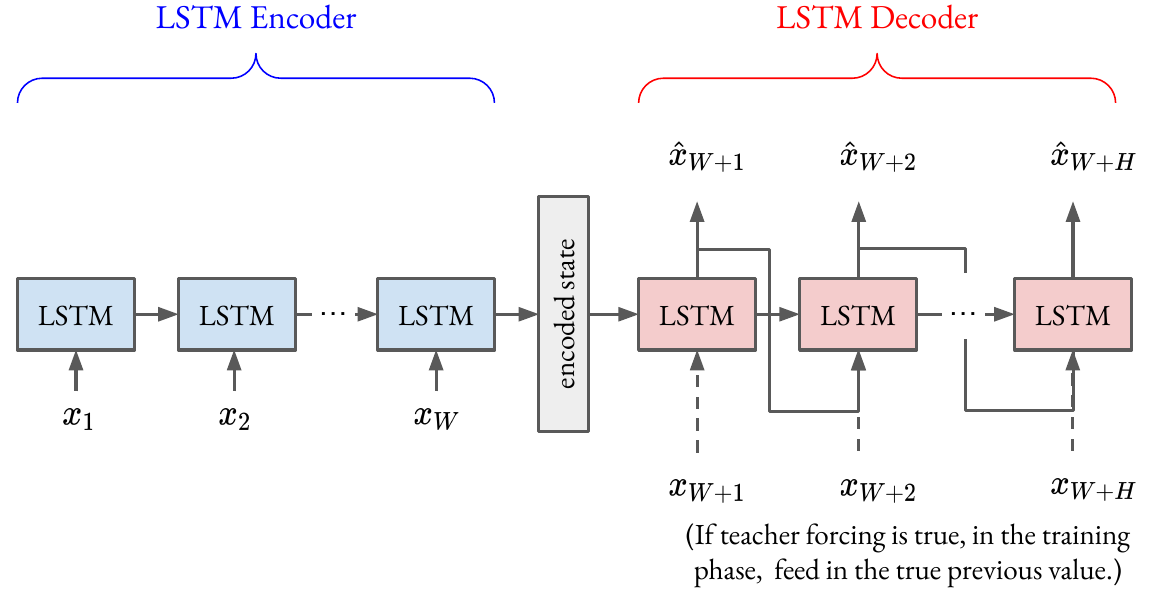}
    \caption{LSTM encoder-decoder architecture.}
    \label{fig:double_rollout}
\end{figure}
\vfill\eject

\section{Additional Results: Average Forecasts}
\begin{figure}[!h]
    \centering
    \includegraphics[width=0.5\columnwidth]{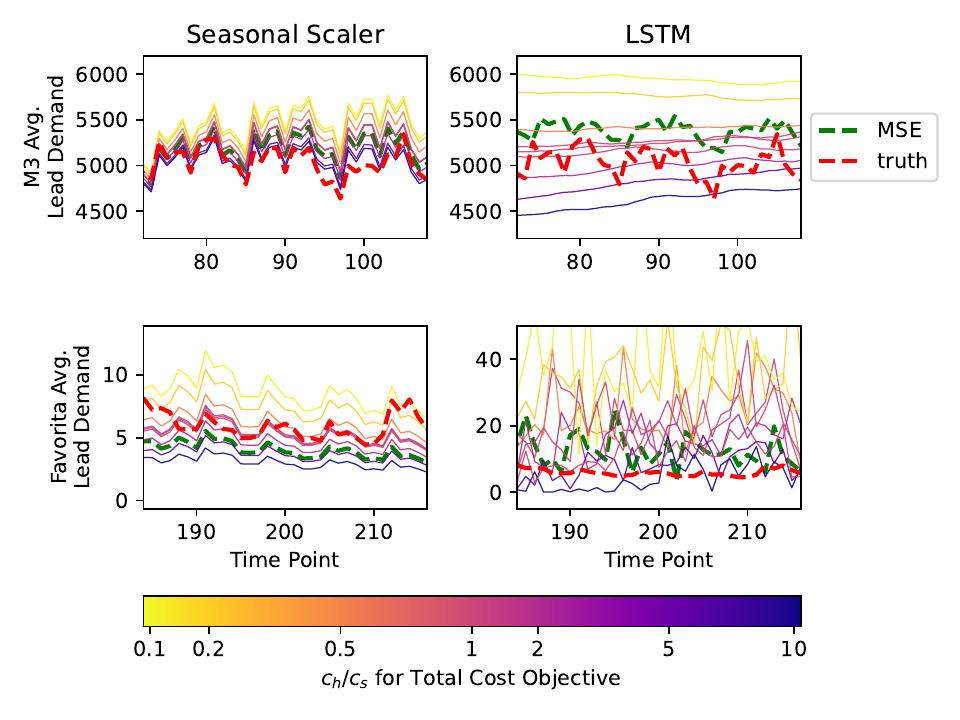}
    \caption{Forecasted and true lead demand, averaged across all series in M3 (first row) and Favorita (second row). True demand is the red dotted line, and forecasts from the MSE objective are the green dotted line. Forecasts from the TC objective are solid lines, for several unit cost tradeoffs---fixing $c_v = 10^{-6}$ in the M3 dataset and $c_v = 10^{-2}$ in the Favorita dataset, ratios of $c_h/c_s$ are indicated by line color.}
    \label{fig:avg_demand}
\end{figure}

Business-aware forecasts are visualized in Figure \ref{fig:avg_demand}.

\section{Additional Results: Relative Improvements with Different Tradeoffs}
\begin{figure}[!h]
    \centering
    \includegraphics[width=0.5\columnwidth]{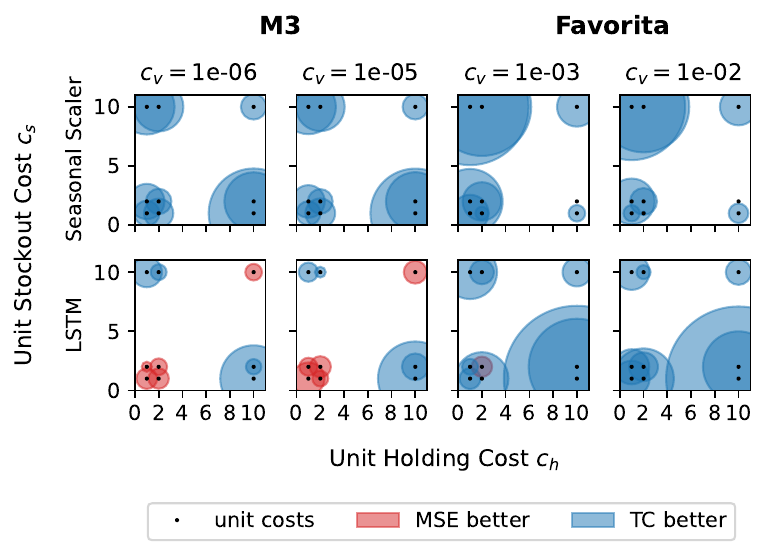}
    \caption{Improvements in total cost when using the TC objective vs. the MSE objective, under various cost tradeoffs. The first two columns correspond to different $c_v$'s on the M3 dataset, and the second two columns correspond to different $c_v$'s on the Favorita dataset. Radius is ten times the relative proportional improvement.}
    \label{fig:both_circles}
\end{figure}

\end{document}